\documentclass[10pt,twocolumn,letterpaper]{article}

\usepackage{cvpr}              %

\usepackage{graphicx}
\usepackage{amsmath}
\usepackage{amssymb}
\usepackage{booktabs}

\usepackage{caption}
\usepackage{float}
\usepackage{booktabs}
\usepackage{graphicx}
\usepackage{tabularx}
\usepackage{patchcmd}
\usepackage{amsmath}
\usepackage{amssymb}
\usepackage{mathtools}

\DeclareMathOperator*{\argmin}{arg\,min}
\DeclarePairedDelimiter{\norm}{\lVert}{\rVert}

\newcommand{\ttilde}{\,\raise.17ex\hbox{$\scriptstyle\sim$}\,}

\usepackage[pagebackref,breaklinks,colorlinks]{hyperref}

\usepackage[capitalize]{cleveref}
\crefname{section}{Sec.}{Secs.}
\Crefname{section}{Section}{Sections}
\Crefname{table}{Table}{Tables}
\crefname{table}{Tab.}{Tabs.}

\def\cvprPaperID{29} %
\def\confName{CVPR}
\def\confYear{2023}

\begin{document}

\title{Happy People -- Image Synthesis as Black-Box Optimization Problem in the Discrete Latent Space of Deep Generative Models}

\author{
Steffen Jung\textsuperscript{1,2} \and
Jan Christian Schwedhelm\textsuperscript{3} \and
Claudia Schillings\textsuperscript{4} \and
 Margret Keuper\textsuperscript{1,2}
\and
\textsuperscript{1} Max Planck Institute for Informatics, Saarland Informatics Campus, \\
\textsuperscript{2} University of Siegen,
\textsuperscript{3} University of Mannheim,
\textsuperscript{4} Free University of Berlin \\
{\tt\small \{steffen.jung,keuper\}@mpi-inf.mpg.de}
}

\twocolumn[{
\maketitle
\vspace{-10mm}
\begin{center}
    \includegraphics[width=0.75\textwidth]{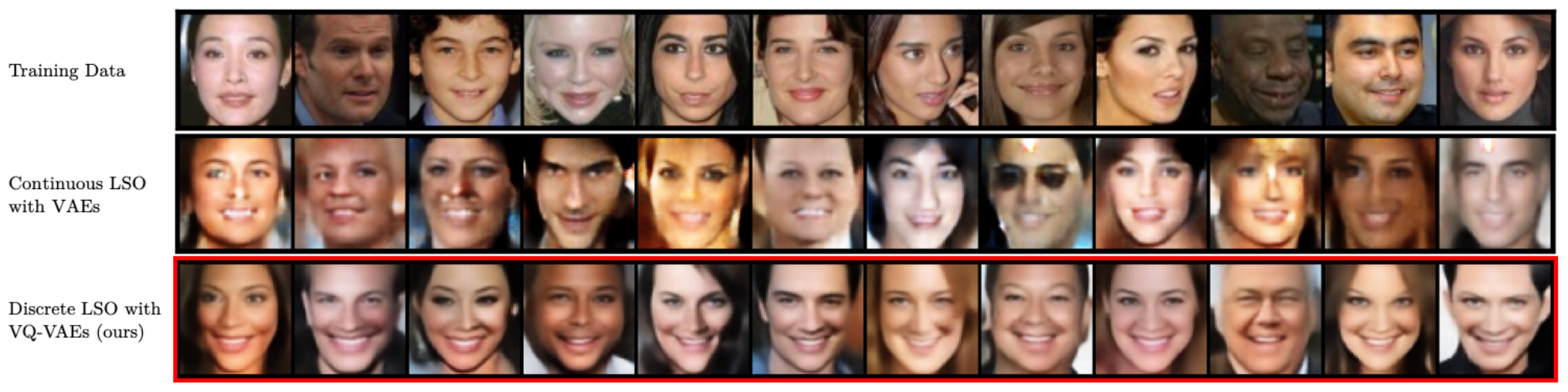}
    \captionof{figure}{In contrast to previous latent space optimization (LSO) approaches (\textbf{middle}), our proposed method is able to synthesize smiling faces with high quality and less artifacts (\textbf{bottom}). Furthermore, the generated images have a significantly higher degree of smiling compared to the best points in the restricted training dataset (\textbf{top}, restricted to smiling degree $\leq2$) that was used to train the models.
    LSO is performed to maximize smiling degree, and FID to the target distribution (unseen  smiling degrees $3-5$) improves from $50.51$ to $41.69$.}
    \label{fig:images}
\end{center}
}]

\begin{abstract}
\vspace{-4mm}
In recent years, optimization in the learned latent space of deep generative models has been successfully applied to black-box optimization problems such as drug design, image generation or neural architecture search. Existing models thereby leverage the ability of neural models to learn the data distribution from a limited amount of samples such that new samples from the distribution can be drawn. %
In this work, we propose a novel image generative approach that optimizes the generated sample with respect to a continuously quantifiable property. While we anticipate absolutely no practically meaningful application for the proposed framework, it is theoretically principled and allows to quickly propose samples at the mere boundary of the training data distribution. 
Specifically, we propose to use tree-based ensemble models as mathematical programs over the discrete latent space of vector quantized VAEs, which can be globally solved.
Subsequent weighted retraining on these queries allows to induce a distribution shift. In lack of a practically relevant problem, we consider a visually appealing application: the generation of happily smiling faces (where the training distribution only contains less happy people) - and show the principled behavior of our approach in terms of improved FID and higher smile degree over baseline approaches.
\end{abstract}

\section{Introduction}
\label{sec:intro}
Many problems in science and engineering can be formulated as optimization of a costly-to-evaluate black-box function over high-dimensional or structured input domains. Notable examples are drug design, which typically requires expensive, time-consuming laboratory experiments for evaluation, neural architecture search, where every potential network solution with variable complexity must be trained and tested, or image synthesis, which can be framed as a black-box optimization problem whose objective function is a human judgment.

In the last few years, \textit{latent space optimization} has been established \cite{DBLP:journals/corr/Gomez-Bombarelli16}, which tackles black-box optimization problems in a two-step procedure: First, a deep generative model (DGM) is trained on the input data, and second, standard optimization methods such as Bayesian optimization (BO) \cite{7352306, brochu2010tutorial} are used in the low-dimensional and continuous latent space learned by the DGM. Despite great successes in application fields such as chemical design \cite{DBLP:journals/corr/Gomez-Bombarelli16, jin2019junction} and automatic machine learning \cite{pmlr-v80-lu18c, eccv}, LSO lacks in performance if the training data of the DGM mainly consists of low-scoring points and the true global
optimum lies far away from this data \cite{tripp2020sampleefficient}. To address these weaknesses, Tripp et al. \cite{tripp2020sampleefficient} have proposed a method to boost the efficiency and performance of LSO by iteratively retraining an encoder-decoder-based DGM (e.g., a Variational Autoencoder (VAE) \cite{kingma2014autoencoding}) on data points queried along the optimization trajectory and weighting those data points according to their objective value. This can be understood as an induced domain shift of a generative model. %
\begin{figure*}[!t]
	\centering
	\includegraphics[width=0.75\textwidth]{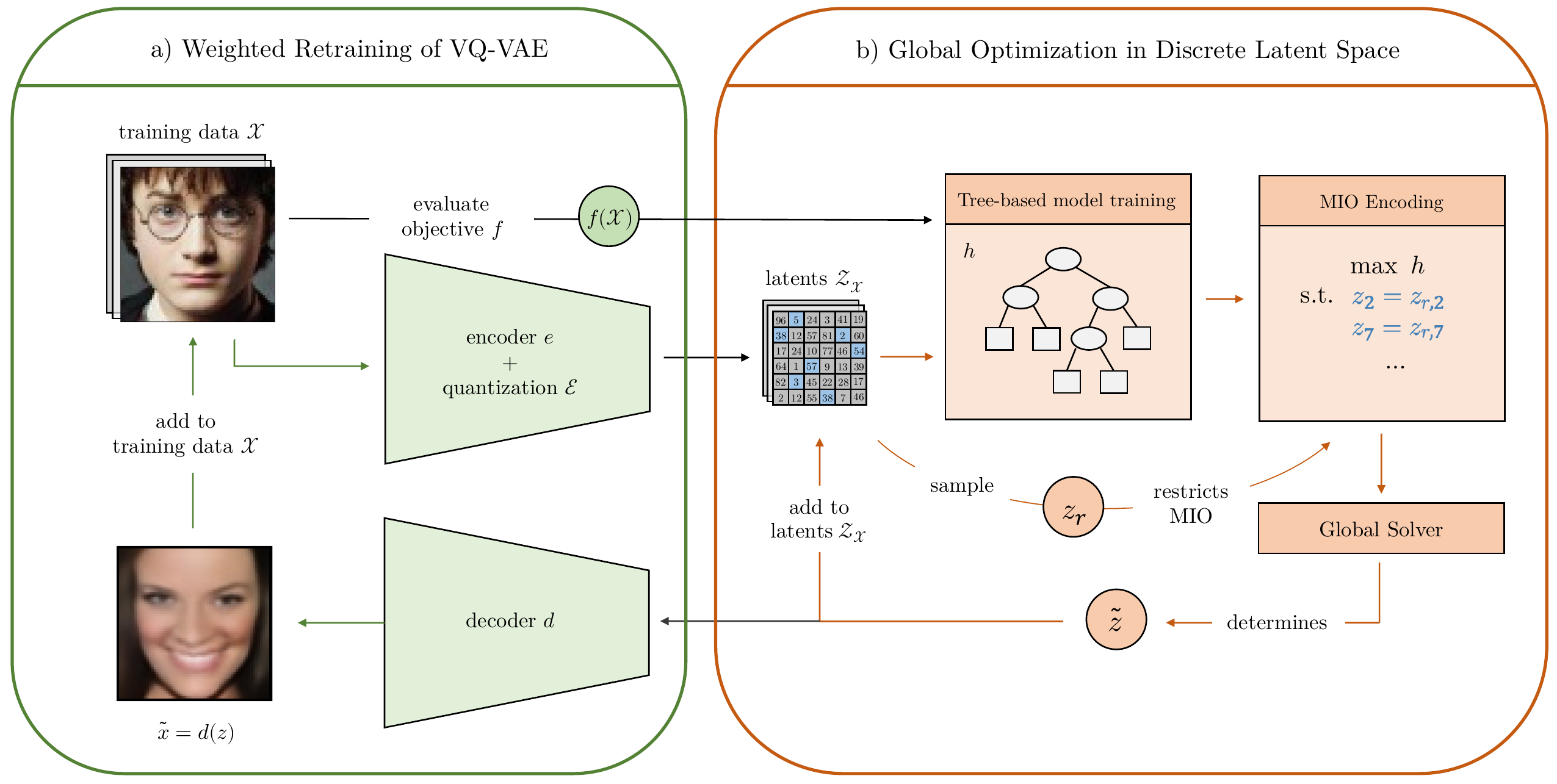}
	\caption{Proposed Framework. \textbf{a)} The encoder and decoder of a VQ-VAE are initially pre-trained and periodically fine-tuned on weighted data $\mathcal{X}$. Images are encoded in a discrete latent space. \textbf{b)} Optimization is performed in the discrete latent space: a tree-based ensemble model is learned from the latent training data representations $\mathcal{Z}_\mathcal{X}$, encoded as a constrained mixed integer optimization problem and globally solved to determine the next (optimal) query point $\tilde{z}$ according to the black-box evaluation function. This procedure is repeated $r$ times before the VQ-VAE is fine-tuned on the queried data points. For this, the corresponding decoded images $\tilde{x}=d(\tilde{z})$ are added to the training data and the weight for each training image is updated.}
	\label{fig:method-overview}
\end{figure*}

By building on \cite{tripp2020sampleefficient}, this paper demonstrates the appliccability of such distribution shifts induced by weighted retraining on black-box optimization problems involving unstructured image data. In particular, we implement \cite{tripp2020sampleefficient} on a generative model involving a discrete latent space, namely \textit{Vector Quantized Variational Autoencoder (VQ-VAE)} \cite{oord2018neural} which constitute very powerful yet simple to train alternatives to VAEs. In contrast to VAEs, VQ-VAEs can generate images at high quality without overly smooth details. %
Classical black-box optimization approaches such as BO are typically based on Gaussian processes \cite{rasmussen2003gaussian} or neural networks \cite{pmlr-v37-snoek15} and thus
assume fully continuous input domains. To allow for targeted optimization within the discrete latent space, this work transfers the LSO process from continuous to categorical input domains by utilizing tree-based ensembles as surrogate objective model in the latent space, encoding their predictions as mathematical optimization programs and solving those programs deterministically using state-of-the-art global solvers, see Figure \ref{fig:method-overview}. As we show in Section \ref{sec:exp}, the proposed framework improves results compared to continuous LSO via standard VAEs, and generates high-quality images that have significantly higher objective function values than the training data.

\section{Tree-based Vector Quantized Generator Optimization}
\label{sec:method}
Although learning representations with continuous variables has been the focus of many previous works \cite{DBLP:journals/corr/ChenDHSSA16, DBLP:journals/corr/DentonGF16, 10.5555/1756006.1953039}, \cite{oord2018neural} has demonstrated that discrete representations learned by VQ-VAEs capture important features of the data with a more natural fit for images. Thus, we provide a solution to transfer the LSO from continuous to discrete latent spaces in order to leverage the expressiveness of VQ-VAEs to generate high-scoring images with good quality. Then, we apply weighted retraining \cite{tripp2020sampleefficient} to induce a distribution shift in the generator distribution. These two steps can be alternated until convergence to train generators with desired target distributions.

\subsection{Discrete latent variables in VQ-VAEs}
\label{subsec:vqvae}
The full VQ-VAE model \cite{oord2018neural} consists of a latent embedding space $\mathcal{E}$, an encoder $e$ and a decoder $d$, where $\mathcal{E}$ contains $K$ learnable vectors $\mathcal{E}_i\in\mathbb{R}^D$, $i\in\{1,\dots,K\}$. %
The shared embedding space $\mathcal{E}$ and the encoder $e$ allow to represent each input image $x\in\mathbb{R}^{H\times W\times 3}$ as a grid $z\in\{1,\dots,K\}^{h\times w}$ of discrete latent variables, where the components
\begin{equation}\notag
    z_{i,j}=\argmin_{l\in\{1,\dots,K\}}
    \norm{e(x)_{i,j,:}-\mathcal{E}_l}
\end{equation}
are obtained by vector quantization of the encoded image $e(x)\in\mathbb{R}^{h\times w\times D}$ w.r.t.~the Euclidean distance. $H\times W$ and $h\times w$ describe the spatial extent of the input and latent representation, respectively.

\subsection{Global Optimization in Discrete Latent Spaces}
\label{subsec:globalopt}

Our method includes optimization in the discrete latent space $z \in \{1,\dots,K\}^{h\times w}$ of a VQ-VAE, which is pre-trained on some input data $\mathcal{X}$.
To this end, a latent objective model $h(\cdot)$ is constructed to approximate the black-box function $f(\cdot)$ at the output of the decoder $d(\cdot)$, i.e., $h(z) \approx f(d(z))$ for all $z\in\{1,\dots,K\}^{h\times w}$. $h$ can be trained by using the encoder $e$ and the learned embedding space $\mathcal{E}$, which together map every $f$-evaluated input image $x\in\mathcal{X}$ to a corresponding discrete latent representation $z = e(x) \in\mathcal{Z}_\mathcal{X}\subset \{1,\dots,K\}^{h\times w}$.

When dealing with categorical feature spaces such as $\{1,\dots,K\}^{h\times w}$, tree-based ensemble models like random forests \cite{breiman2001random} or gradient-boosting trees \cite{10.1214/aos/1013203451} are popular choices for $h$, since they naturally support various data types. Thus, we train a decision tree ensemble model to predict the objective value of a given image sample from its discrete latent representation. 
Interestingly, \cite{entmoot2021} propose an intricate approach that allows to encode the trained tree-based model with a mixed-integer optimization (MIO) formulation %
which allows the optimization of the discrete latent code w.r.t.~the objective value, the tree-based model has been trained to predict. Please refer to \cite{entmoot2021} for the theoretical proof. By solving the resulting MIO program deterministically using a global solver \cite{kraft1988software}, $h$ can be optimized to determine the next latent query point $\tilde{z}$. 
During latent optimization, the solution $\tilde{z}$ has to be restricted to stay sufficiently close to latent representations of the training data. Otherwise, without any constraints on the latent variables, the decoded version $\tilde{x}=d(\tilde{z})$ of the query point $\tilde{z}$ most likely has bad quality for the following reason: values of those latent variables not having an impact on image regions that determine the objective function value can arbitrarily be chosen in the optimization process, which may lead to highly distorted image features. %
We address this problem as follows: First, a single training sample $x_r\in\mathcal{X}$ is randomly drawn and mapped to its latent representation $z_r$. Second, only those $t\in\mathbb{N}$ latent variables having the highest feature importances under the trained tree-based model are free to be globally optimized, while the remaining $hw-t$ variables are fixed by being set to the respective entries of $z_r$. 
Taking into account the optimization result $\tilde{z}$ together with the corresponding objective function evaluation $f(\tilde{x})$ of its decoded version $\tilde{x}=d(\tilde{z})$, $h$ is refit after every iteration.

\subsection{Weighted Retraining}
Subsection \ref{subsec:globalopt} introduces a technique to perform LSO in discrete latent spaces. However, identified in \cite{tripp2020sampleefficient}, the underlying DGM does not necessarily learn a latent space that is amenable to efficient optimization of the objective function, especially in cases where the global optimum is far away from the training data.
To resolve this issue, we follow \cite{tripp2020sampleefficient} and weight the training data points $\mathcal{X}=\{x_i\}_{i=1}^N$ according to their objective values $\{f(x_i)\}_{i=1}^N$: the higher a value $f(x)$, the more probability mass the training distribution should place on the corresponding input point $x$. The weighting scheme requires assigning a weight $w(x_i)\in\mathbb{R}$ with $\sum_{i=1}^N w(x_i)=1$ for all $x_i\in \mathcal{X}$ using some weight function $w$. In this work, we adopt the rank-based function introduced in \cite{tripp2020sampleefficient}. %

To propagate information on new points acquired during the iterative LSO process to the VQ-VAE, where it could potentially help to uncover new promising regions that an optimization algorithm can exploit, the VQ-VAE is periodically fine-tuned after every $r\in\mathbb{N}$ LSO iterations.

\section{Experiments}
\label{sec:exp}
Here, we provide an empirical evaluation of the proposed discrete LSO with weighted retraining, as introduced in Section \ref{sec:method}. First, we define a challenging optimization task using the famous face dataset CelebA \cite{liu2015faceattributes}.
Based on this image synthesis task, we evaluate our methods’s ability to compete with continuous LSO via standard VAEs  \cite{tripp2020sampleefficient}. For the underlying VQ-VAE, we use the CNN encoder-decoder architecture introduced in \cite{oord2018neural}. Please refer to the appendix for more details.

\label{sec:prep}

\textbf{Optimization task.}
We employ CelebA $64\times 64$ for an image synthesis task which can be viewed as black-box optimization problem. Our goal is to generate smiling faces by optimizing for the respective attribute degree in the space of colored $64 \times 64$ face images. To represent the smiling attribute on a continuous scale, we make use of the extended CelebA-Dialog dataset \cite{jiang2021talkedit}, which includes fine-grained labels for five selected CelebA attributes that cannot be accurately described by binary labels. In particular, the smiling attribute is divided into six levels ($0-5$) that describe the degree in ascending order. For unseen face images, we estimate the degree of smiling by using a probability-weighted average $\hat{f}_\text{smile}\colon\mathbb{R}^{64\times 64\times 3}\rightarrow [0,5]$ of class predictions coming from the ResNet-$50$ \cite{he2015deep} classifier used in the official implementation of \cite{jiang2021talkedit}, which is pre-trained on CelebA-Dialog. %

We discard points with high objective values from the training data to make the problem more challenging and to represent the situation where the optimum (degree $5$) lies far outside the training distribution. Specifically, \emph{only images with a smiling degree $\leq 2$ are kept in the training set.}

\begin{figure*}[!t]
	\centering
	\includegraphics[width=0.8\textwidth]{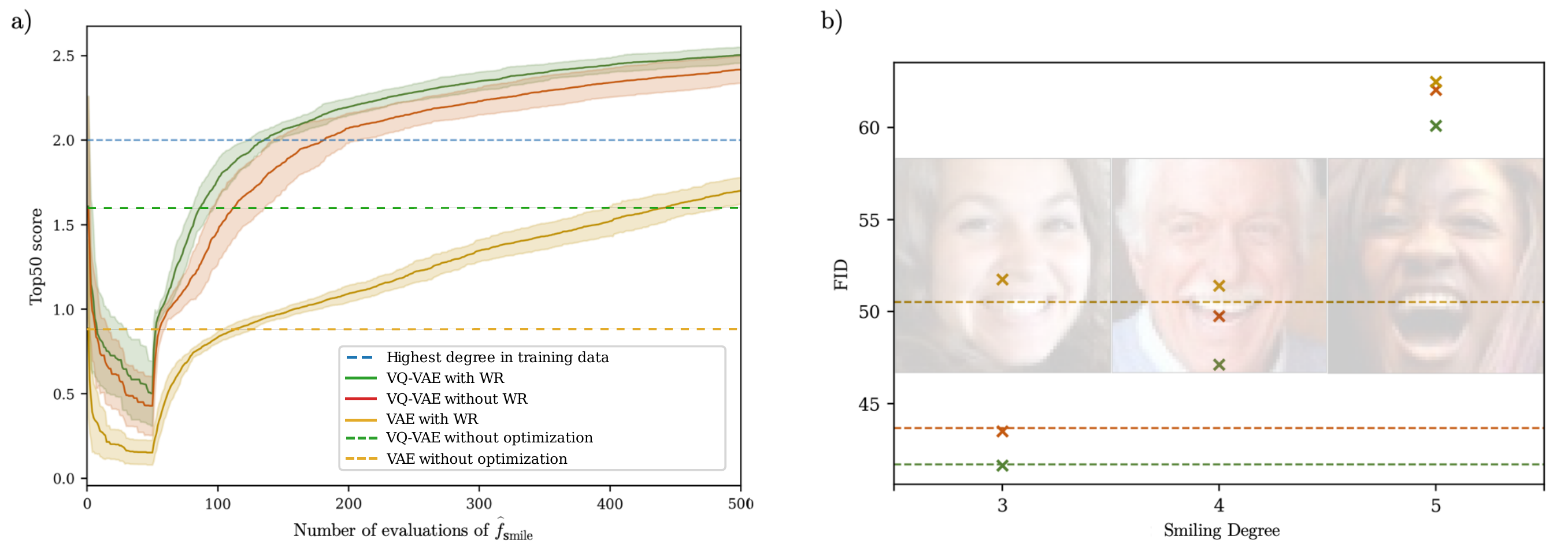}
	\caption{Quantitative comparison between discrete and continuous LSO variants: VQ-VAE with weighted retraining (\textbf{green}), VQ-VAE without weighted retraining (\textbf{red}), and VAE with weighted retraining (\textbf{yellow}). Evaluation metrics: \textbf{a)} Top$50$ score function, where the blue dashed line depicts the best smiling degree in the training data.
 VQ-VAE as well as weighted retraining improve the degree of smiling substantially.
 \textbf{b)} FID scores, where the dashed lines show the respective scores of testing against the subset of CelebA images having an unseen smiling degree ($3-5$), and the markers show testing against each training data subset with the corresponding unseen smiling degree ($3$, $4$, and $5$; 
 background images depict training data samples for these smiling degrees).
 In all cases, VQ-VAE with weighted retraining achieves the best result.
}
	\label{fig:metrics}
\end{figure*}
\textbf{Implementation details.}
Throughout this work, we assume a query budget of $500$ and a retrain frequency of $r=5$.
For the underlying VQ-VAE, we use the CNN encoder-decoder architecture as in \cite{oord2018neural}. %
We assume a discrete latent space of size $8\times 8$ ($64$ latent variables in total). Moreover, we consider $K = 256$ embedding vectors with a dimensionality of $D = 64$ each.
The tree-based model $h$ is chosen to be an ensemble of $800$ gradient-boosted regression tree and an interaction depth of $2$, following \cite{entmoot2021}.
The minimum number of samples in one leaf equals $20$, and the maximum number of leaves per tree is set to $5$. 

Continuous LSO with VAEs is implemented by a neural network as latent objective model, as in \cite{pmlr-v37-snoek15}. %
Motivated by \cite{Ghoshetal19}, local optimization in the continuous latent space is constrained using ex-post density estimation via Gaussian Mixture Models.
Optimization is carried out using the SLSQP algorithm \cite{kraft1988software}.

\textbf{Evaluation metrics.} Following common practice for BO \cite{7352306, tripp2020sampleefficient},  we show the worst of the $10$ and $50$ best novel smiling degree predictions obtained up until query $m=1,\dots,500$, which is denoted as Top$10$ and Top$50$ score function, respectively. To ensure statistical significance, the mean $\pm$ one standard deviation across $20$ runs with different random seeds is reported. Furthermore, we use FID scores \cite{heusel2018gans} as quantitative assessment of the quality of generated images. Since our goal is to find faces that have a higher smiling degree than the best point in the training data, we compute FID scores between all $10,000$ generated images from the $20$ runs and the subset of CelebA faces having a smiling degree between $3$ and $5$.

\textbf{Results on smiling face synthesis. }
Figure \ref{fig:metrics} presents quantitative comparison between our approach and continuous LSO via standard VAEs. Corresponding visual results are shown in Figure \ref{fig:images}.  %
Even if weighted retraining is applied, the Top$50$ scores resulting from LSO with VAEs are strictly below the value $2.0$ that corresponds to the best training images (mean final score: $1.70$). Even without weighted retraining of the VQ-VAE, our method outperforms VAEs by a large margin (mean final score: $2.42$). Weighted retraining further improves the final Top$50$ score from $2.42$ to $2.50$ on average.

Moreover, the FID results show that our method successfully leverages the expressiveness of VQ-VAEs to generate images with higher quality compared to standard VAEs. The FID score can be decreased by $21\%$, from $50.51$ to $41.69$. Again, weighted retraining hereby has a positive effect and leads to a significant improvement of $5\%$. In addition, we tested against three separate subsets containing all images having a smiling degree of $3$, $4$, and $5$, respectively, to detect potential biases among the generated faces. However, for all considered LSO variants, the results are as expected: higher smiling degrees in the target distribution lead to higher FID scores. Overall, the level of FID scores is relatively high, which can likely be attributed to the following reason: the target distribution (degrees $3-5$) is different from the distribution the VQ-VAE is pre-trained on (degrees $0-2$), which prevents direct comparison to other generative models that are explicitly designed and trained to generate samples that resemble the given training data.

\section{Conclusion}
We propose a method for efficient black-box optimization in the discrete latent space of VQ-VAEs, which combines (i) choosing a tree-based ensemble as latent objective model, (ii) encoding its predictions as an MIO problem that is solved globally to determine the next query point and (iii) iteratively fine-tuning the underlying VQ-VAE on weighted data. With the challenging task of generating smiling faces that are not contained in the training distribution, we demonstrate that our method notably outperforms continuous LSO with VAEs in terms of both image quality and optimization of the objective function. To the best of our knowledge, this is the first work that successfully applies LSO in discrete latent spaces on image synthesis tasks. %

{\small
\bibliographystyle{ieee_fullname}
\bibliography{egbib}
}

\appendix
\onecolumn
\section{Appendix}
\subsection{Architectures and Implementation} For the underlying VQ-VAE, we use the CNN encoder-decoder architecture as introduced in \cite{oord2018neural}. %
We assume a discrete latent space of size $8\times 8$, which results in $64$ latent variables in total. Moreover, we consider $K = 256$ embedding vectors with a dimensionality of $D = 64$ each. The tree-based model $h$ is chosen to be a gradient-boosting regression tree with an ensemble of $800$ decision trees and an interaction depth of $2$, following \cite{entmoot2021}. Furthermore, the minimum number of samples in one leaf equals $20$, and the maximum number of leaves per tree is set to $5$. 
During LSO, we want to preserve as much image quality as possible. Therefore, we will not optimize over all $64$ latent variables but only over those that are most important for the objective value we are optimizing for. Thus, we leverage the feature importance estimated by the decision tree ensemble to determine latent variables to optimize. An ablation of the number of latent variables used for optimization is given in Section \ref{abl}.

Continuous LSO with VAEs is implemented by a neural network as latent objective model, as in \cite{pmlr-v37-snoek15}. %
Motivated by \cite{Ghoshetal19}, local optimization in the continuous latent space is constrained using ex-post density estimation via Gaussian Mixture Models, and new query points are restricted to stay sufficiently close to the estimated distribution. Optimization is carried out using the SLSQP algorithm \cite{kraft1988software}.

\subsection{Ablation study}
\label{abl}
During LSO, we want to preserve as much image quality as possible. Therefore, we will not optimize over all $64$ latent variables but only over those that are most important for the objective value we are optimizing for. Here, we measure the impact of the number $t$ of free LSO variables on the performance of our proposed method.
For this purpose, we test $4$, $8$, and $16$ free variables for $20$ optimization runs each. FID scores as well as final Top$10$ and Top$50$ scores after $500$ optimization iterations are reported in Table \ref{tab:ablation}.
	\begin{table}[!h]
	\centering
	\small
	\captionsetup{font=small}
	\captionsetup{justification=centering, skip=5pt}
	\begin{tabular}{r c c c}				
		\toprule
		\textbf{$t$} &  \textbf{FID} & \textbf{Top$10$} & \textbf{Top$50$} \\
		\toprule

		 $4$ & $42.69$ & $2.72\pm 0.07$ & $2.42\pm 0.04$  \\
		 $8$ & $41.69$ & $2.85\pm 0.07$ & $2.50\pm 0.05$ \\
		 $16$ & $43.70$ & $2.88\pm0.06$ & $2.58\pm 0.02$ \\
		\bottomrule
	\end{tabular}\\
	\caption{Ablation study results. Top$10$ and Top$50$ scores are reported as mean final score $\pm$ one standard deviation after reaching the query budget of $500$.}
	\label{tab:ablation}
\end{table}

As we can observe, choosing $t=8$ hits a sweet spot: while generated faces have significantly lower quality if $t$ is increased (FID: $41.69$ vs.~$43.70$), a reduction of $t$ leads to lower smiling degrees (Top$10$: $2.85$ vs.~$2.72$). We set $t$ to 8 for all experiments.
\subsection{Details on VQ-VAE}
\begin{table}[H]
	\centering
	\scriptsize
	\captionsetup{font=small}
	\captionsetup{justification=centering, skip=5pt}
		\begin{tabular}{r l c c c c c c}				
			\toprule
			\textbf{Stage} &  \textbf{Layer} &  \textbf{Input Shape} &  \textbf{Output Shape} & \textbf{$\#$ Filters} & \textbf{Kernel Size} & \textbf{Padding} & \textbf{Stride} \\
			\quad &  \quad &  $(H_0, W_0, D_0)$ &  $(H_1, W_1, D_1)$ & $K$ & $F$ & $P$ & $S$ \\
			\toprule
			1 &  Conv &  $(64,64,3)$ &  $(32,32,64)$ & $64$ & $4$ & $1$ & $2$ \\
			&  Leaky ReLu &  &   & &  &  &  \\
			2 &  Conv &  $(32,32,64)$ &  $(16,16,128)$ & $128$ & $4$ & $1$ & $2$ \\
			& Leaky ReLu &  &   & &  &  &  \\
			3 &  Conv &  $(16,16,128)$ &  $(8,8,256)$ & $256$ & $4$ & $1$ & $2$ \\
			&  Leaky ReLu &  &   & &  &  &  \\
			4 &  Conv &  $(8,8,256)$ &  $(8,8,256)$ & $256$ & $3$ & $1$ & $1$ \\
			&  Leaky ReLu &  &   & &  &  &  \\
			\midrule
			5 &  Residual Block &  $(8,8,256)$ &  $(8,8,256)$ &  &  &  &  \\
			6 &  Residual Block &  $(8,8,256)$ &  $(8,8,256)$ &  &  &  &  \\
			\midrule
			7 &  Conv &  $(8,8,256)$ &  $(8,8,64)$ & $64$ & $1$ & $0$ & $1$ \\
			&  Leaky ReLu &  &   & &  &  &  \\
			\bottomrule
	\end{tabular}\\
	\caption{Encoder architecture of VQ-VAE}\label{tab:encoder-celeba-vqvae}
\end{table}
\begin{table}[H]
	\centering
	\scriptsize
	\captionsetup{font=small}
	\captionsetup{justification=centering, skip=5pt}
		\begin{tabular}{r l c c c c c c c}				
			\toprule
			\textbf{Stage} &  \textbf{Layer} &  \textbf{Input Shape} &  \textbf{Output Shape} & \textbf{$\#$ Filters} & \textbf{Kernel Size} & \textbf{Padding} & \textbf{Stride} & \textbf{O-Padding} \\
			\quad &  \quad &  $(H_0, W_0, D_0)$ &  $(H_1, W_1, D_1)$ & $K$ & $F$ & $P$ & $S$ & $O$ \\
			\toprule
			1 &  Conv &  $(8,8,64)$ &  $(8,8,256)$ & $256$  & $3$ & $1$ & $1$ &\\
			&  Leaky ReLu &  &   & &  &  & & \\
			\midrule 
			2 &  Residual Block &  $(8,8,256)$ &  $(8,8,256)$ &  &  &  & &  \\
			3 &  Residual Block &  $(8,8,256)$ &  $(8,8,256)$ &  &  &  & &  \\
			\midrule
			4 &  ConvT &  $(8,8,256)$ &  $(16,16,128)$ & $128$ & $4$ & $1$ & $2$ & $0$\\
			&  Leaky ReLu &  &   & &  &  & & \\
			5 &  ConvT &  $(16,16,128)$ &  $(32,32,64)$ & $64$ & $4$ & $1$ & $2$ & $0$\\
			&  Leaky ReLu &  &   & &  &  & & \\
			6 &  ConvT &  $(32,32,64)$ &  $(64,64,3)$ & $3$ & $4$ & $1$ & $2$ & $0$\\
			\bottomrule
	\end{tabular}\\
	\caption{Decoder architecture of VQ-VAE}\label{tab:decoder-celeba-vqvae}
\end{table}
\begin{table}[H]
	\centering
	\small
	\captionsetup{font=small}
	\captionsetup{justification=centering, skip=5pt}
	\begin{tabular}{l c}				
		\toprule
		\textbf{Hyperparameter} &  \textbf{Value}\\
		\toprule
		Latent Dimension & $64$ \\
		Embedding Dimension $D$ & $64$ \\
		$\#$ Categories $K$ & $256$ \\
		Batch Size & $128$ \\
		$\#$ Epochs & $70$ \\
		Optimizer & Adam \\
		Learning Rate & $0.001$ \\
		\midrule
		$\beta$ & $0.25$ \\
		\bottomrule
	\end{tabular}\\
	\caption{Training hyperparameters of VQ-VAE}
	\label{tab:train_params_celeba_vqvae}
\end{table}
\pagebreak

\subsection{Details on VAE}
	\begin{table}[H]
	\centering
	\scriptsize
	\captionsetup{font=small}
	\captionsetup{justification=centering, skip=5pt}
		\begin{tabular}{r l c c c c c c}				
			\toprule
			\textbf{Stage} &  \textbf{Layer} &  \textbf{Input Shape} &  \textbf{Output Shape} & \textbf{$\#$ Filters} & \textbf{Kernel Size} & \textbf{Padding} & \textbf{Stride} \\
			\quad &  \quad &  $(H_0, W_0, D_0)$ &  $(H_1, W_1, D_1)$ & $K$ & $F$ & $P$ & $S$ \\
			\toprule
			1 &  Conv &  $(64,64,3)$ &  $(32,32,32)$ & $32$ & $3$ & $1$ & $2$ \\
			&  Leaky ReLu &  &   & &  &  &  \\
			&  BatchNorm &  &   & &  &  &  \\
			2 &  Conv &  $(32,32,32)$ &  $(16,16,64)$ & $64$ & $3$ & $1$ & $2$ \\
			&  Leaky ReLu &  &   & &  &  &  \\
			&  BatchNorm &  &   & &  &  &  \\
			3 &  Conv &  $(16,16,64)$ &  $(8,8,128)$ & $128$ & $3$ & $1$ & $2$ \\
			&  Leaky ReLu &  &   & &  &  &  \\
			&  BatchNorm &  &   & &  &  &  \\
			4 &  Conv &  $(8,8,128)$ &  $(4,4,256)$ & $256$ & $3$ & $1$ & $2$ \\
			&  Leaky ReLu &  &   & &  &  &  \\
			&  BatchNorm &  &   & &  &  &  \\
			5 &  Conv &  $(4,4,256)$ &  $(2,2,512)$ & $512$ & $3$ & $1$ & $2$ \\
			&  Leaky ReLu &  &   & &  &  &  \\
			&  BatchNorm &  &   & &  &  &  \\
			\midrule
			6&  Flatten & $(2,2,512)$ & $2048$  & &  &  &  \\
			&  Dense &  $2048$ &  $128$ &  &  &  &  \\
			\bottomrule
	\end{tabular}\\
	\caption{Encoder architecture of VAE}\label{tab:encoder-celeba-vae}
\end{table}
\begin{table}[!htbp]
	\centering
	\scriptsize
	\captionsetup{font=small}
	\captionsetup{justification=centering, skip=5pt}
		\begin{tabular}{r l c c c c c c c}				
			\toprule
			\textbf{Stage} &  \textbf{Layer} &  \textbf{Input Shape} &  \textbf{Output Shape} & \textbf{$\#$ Filters} & \textbf{Kernel Size} & \textbf{Padding} & \textbf{Stride} & \textbf{O-Padding} \\
			\quad &  \quad &  $(H_0, W_0, D_0)$ &  $(H_1, W_1, D_1)$ & $K$ & $F$ & $P$ & $S$ & $O$ \\
			\toprule
			1 &  Dense &  $64$ &  $2048$ &  &  & & &\\
			&  Leaky ReLu &  &   & &  &  & & \\
			&  BatchNorm &  &   & &  &  & & \\
			2 &  Unflatten &  $2048$ &  $(2,2,512)$ &  & & & \\
			&  ConvT &  $(2,2,512)$ &  $(4,4,256)$ & $256$ & $3$ & $1$ & $2$ & $1$\\
			&  Leaky ReLu &  &   & &  &  & & \\
			&  BatchNorm &  &   & &  &  & & \\
			3 &  ConvT &  $(4,4,256)$ &  $(8,8,128)$ & $128$ & $3$ & $1$ & $2$ & $1$\\
			&  Leaky ReLu &  &   & &  &  & & \\
			&  BatchNorm &  &   & &  &  & & \\
			4 &  ConvT &  $(8,8,128)$ &  $(16,16,64)$ & $64$ & $3$ & $1$ & $2$ & $1$\\
			&  Leaky ReLu &  &   & &  &  & & \\
			&  BatchNorm &  &   & &  &  & & \\
			5 &  ConvT &  $(16,16,64)$ &  $(32,32,32)$ & $32$ & $3$ & $1$ & $2$ & $1$\\
			&  Leaky ReLu &  &   & &  &  & & \\
			&  BatchNorm &  &   & &  &  & & \\
			6 &  ConvT &  $(32,32,32)$ &  $(64,64,32)$ & $32$ & $3$ & $1$ & $2$ & $1$\\
			&  Leaky ReLu &  &   & &  &  & & \\
			&  BatchNorm &  &   & &  &  & & \\
			\midrule
			7 &  Conv &  $(64,64,32)$ &  $(64,64,3)$ & $3$ & $3$ & $1$ & $1$ &  \\
			
			\bottomrule
	\end{tabular}\\
	\caption{Decoder architecture of VAE}\label{tab:decoder-celeba-vae}
\end{table}
\begin{table}[!htbp]
	\centering
	\small
	\captionsetup{font=small}
	\captionsetup{justification=centering, skip=5pt}
	\begin{tabular}{l c}				
		\toprule
		\textbf{Hyperparameter} &  \textbf{Value}\\
		\toprule
		Latent Dimension & $64$ \\
		Batch Size & $64$ \\
		$\#$ Epochs & $30$ \\
		Optimizer & Adam \\
		Learning Rate & $0.0001$ \\
		\midrule
		$\beta_{\text{start}}$ & $10^{-6}$ \\
		$\beta_{\text{final}}$ & $0.215$ \\
		$\beta$ Warm-up & $5000$ \\
		$\beta$ Step & $1.1$ \\
		$\beta$ Step Frequency & $50$ \\
		\bottomrule
	\end{tabular}\\
	\caption{Training hyperparameters of VAE}
	\label{tab:train_params_celeba_vae}
\end{table}

\subsection{Details on dataset}
We focus on the facial part of CelebA images by using pre-trained Multitask Cascaded Convolutional Networks (MTCNN) \cite{mtcnn} to localize faces in the input plane, and cropping images according to the predicted bounding box. Furthermore, the data is resized to $64\times 64$ via resampling using pixel area relation.

\end{document}